\documentclass[runningheads]{llncs}
\usepackage{graphicx}
\usepackage{algorithm}
\usepackage[noend]{algcompatible}
\newcommand{\algorithmicinput}{\textbf{input}}
\newcommand{\INPUT}{\item[\algorithmicinput]}
\usepackage{amsfonts}

\begin{document}
\title{VAESim: A probabilistic approach for self-supervised prototype discovery}
\author{Matteo Ferrante\inst{1} \and
Tommaso Boccato\inst{1} \and
Simeon Spasov \inst{3} \and
Andrea Duggento \inst{1} \and
Nicola Toschi\inst{1,2}}
\authorrunning{Ferrante et al.}
\institute{Department of Biomedicine and Prevention, University of Rome Tor Vergata (IT) \and Martinos Center For Biomedical Imaging, MGH and Harvard Medical School (USA) \and Department of Computer Science and Technology, University of Cambridge (UK) }
\maketitle              % typeset the header of the contribution

\begin{abstract}
In medicine, curated image datasets often employ discrete labels to describe what is known to be continuous spectrum of healthy to pathological conditions, such as e.g. the Alzheimer’s Disease Continuum or other areas where the image plays a pivotal point in diagnosis.
We propose am architecture for image stratification based on a conditional variational autoencoder. Our framework, VAESim, leverages a continuous latent space to represent the continuum of disorders and finds clusters during training, which can then be used for image/patient stratification. The core of the method learns a set of prototypical vectors, each associated with a cluster. First, we perform a soft assignment of each data sample to the clusters. Then, we reconstruct the sample based on a similarity measure between the sample embedding and the prototypical vectors of the clusters. To update the prototypical embeddings, we use an exponential moving average of the most similar representations between actual prototypes and samples in the batch size.
We test our approach the MNIST-handwritten digit dataset and on a medical benchmark dataset called PneumoniaMNIST. We demonstrate that our method outperforms baselines in terms of kNN accuracy measured on a classification task against a standard VAE (up to $+15\%$ improvement in performance) in both datasets, and also performs at par with classification models trained in a fully supervised way. We also demonstrate how our model ourperforms current, end-to-end models for unsupervised stratification.

\end{abstract}

\section{Introduction}

In medicine, large unlabeled image datasets vastly outnumber curated datasets. Furthermore, when labels are available, discrete labels are often employed to describe a continuous spectrum of conditions. There is a lot of interest in developing unsupervised methods to uncover the hidden, or latent, structure of these disorders, in order to identifying disease subtypes or novel classess of disease (see [1] for a review). Unsupervised learning, however, is a challenging task to solve even in domains where large and curated datasets are available. Many diverse approaches have been recently proposed, however they commonly rely on extremely large amounts of data. I norder to render such approaches more practical, a possible solution could be the development of data-efficient unsupervised methods or an increased reliance on self-supervised or unsupervised pretraining followed by supervised fine-tuning, which would drastically reduce the need for labeled samples. In thei context, a number of papers have tackled the problem of clustering using deep learning architectures that find a nonlinear mapping from the input space to the feature space, where clustering is performed (see “Related Work” section).
In this work, we propose an unsupervised deep learning probabilistic approach that projects the inputs to a learned latent space with “prototype” points, such that the low-dimensional representations of the inputs are useful both for their reconstructions and for subsequent tasks such as classifications. Each “prototype” point is associated with a cluster in the latent space. During reconstruction, we condition the decoder on both a) the sample latent embedding and b) a soft cluster assignment, which relies on the similarity of the sample embedding to all prototype vectors. We assume that downstream tasks, such as e.g. classification, are simplified by this conditioning, which provides more context on the relative position of the sample embedding in the latent space. In this context, the main challenge is defining a strategy to learn these prototype vectors, which should act represent subsets that cluster together in the latent space.
The objective of this work is to augment the VAE framework to also learn prototype vectors, hence obtaining a better organization of the latent space. The prototypes are learned using a momentum update during training and are stored in the memory of the model. This conditional information about similarity during decoding should carry all the information needed for reconstructions and implicitly impose a structure in the latent space. We demonstrate that this approach is more effective than two-step approaches that first train a vanilla VAE and then cluster its latent space, for example, by using KMeans. We show that our end-to-end approach outperforms these and other deep clustering baselines on the MNIST handwritten digit dataset  \cite{mnist}as a benchmark and on the Pneumonia dataset from the MedMNIST collection \cite{medmnistv2} as an example in the field of biomedical images. These performance differences are quantified in terms of different metrics, related to both the alignment between the label clusters and the classification obtained using kNN, or linear classification to measure the capacity of the latent space to capture the fundamental characteristics of the encoded samples.

\section{Related Work}
General trends in unsupervised and self-supervised deep tend to be dominated by large networks trained over hundreds of thousands of images. Recent examples include student-teacher approaches such as DINO  \cite{dino}, based on knowledge distillation without the need for labels, and contrastive-based approaches such as \cite{simclr,simclr2,moco}, where models are trained through a contrastive loss and a similarity metric to position similar samples close in the latent space and different ones very far apart and make heavy use of different augmentations to generate positive pairs. However, due to the intrinsic properties of medical image datasets, these methods poorly translate to the medical context. Knowledge distillation approaches such as DINO rely on the use of data-hungry architectures like vision transformers \cite{vit} or ResNet50 \cite{resnet}], whose power can be harnessed only when “big data” is available. Contrastive-based approaches can tackle label scarcity by employing augmentations, but in medical applications, geometrical augmentations such as mirrors, zooms, and color distribution shifts cannot capture the full pathology diversity. 
Furthermore, approaches like SimCLR \cite{simclr} need a very large batch size to consider negative sampling, resulting in high computational demands that can only be met with nonstandard hardware. This problem could be partially solved using more efficient methods such as MOCO  \cite{moco}where a momentum encoder is used for negative sampling storing low-dimensional representations of the precedent batches. Even if all these approaches are appealing and promising when tested on large-scale datasets, their direct translation to data from the medical domain is still not straightforward.
Smaller models could be useful when data are scarce and are mainly built on self-supervised pretraining using autoencoders or variational autoencoders. In \cite{deepclustervariationautoenc} ], the authors trained a variational autoencoder using information from a Gaussian mixture model (GMM) employed as an unsupervised clustering algorithm. Including this information in the loss function, they were able to drive the model to learn a well-separated latent space. In  \cite{vade} ], the authors replaced the VAE prior with a GMM-based prior enforcing the same type of separation in a more direct way. In \cite{deepkmeans}.
], a differentiable version of the K-Means algorithm based on the softmax function (instead of argmax) is proposed and embedded in a neural network. All these methods propose different solutions to the clustering problem by exploiting classification with soft labels or even reconstruction of samples, in most of cases using benchmark datasets as proof-of-concept demonstrations. In general, one of the main open issues is how to inject information about cluster labels, which are usually obtained using nondifferentiable functions, into the model. In our work, we combine soft labels obtained by tempered softmax similarity to inform the decoder about clustering and use nondifferentiable functions to update the prototype values.
Other unsupervised approaches extended those and other ideas to the medical field. Examples are the Mixture of Experts (MoE)  \cite{moe}, where a clustering network proposes a cluster and a mixture of experts learn how to reconstruct the data related to single-cell data, or \cite{CNN-Kmeans-medical} where a CNN is combined with a KMeans approach to classifying an image dataset with different modalities. Finally, a translation of contrastive learning, which also uses multi-instance learning, has achieved results comparable to fully supervised state-of-art algorithms \cite{micle}.
Again, however, in many clinical/medical domains, these strategies might be suboptimal because they rely on nondifferentiable methods, and require a large amount of data or domain-specific knowledge necessary to incorporate biologically/medically aware data augmentations.

\section{Materials and methods}

\begin{figure}[h]
    \centering
    \includegraphics[width=0.6\linewidth]{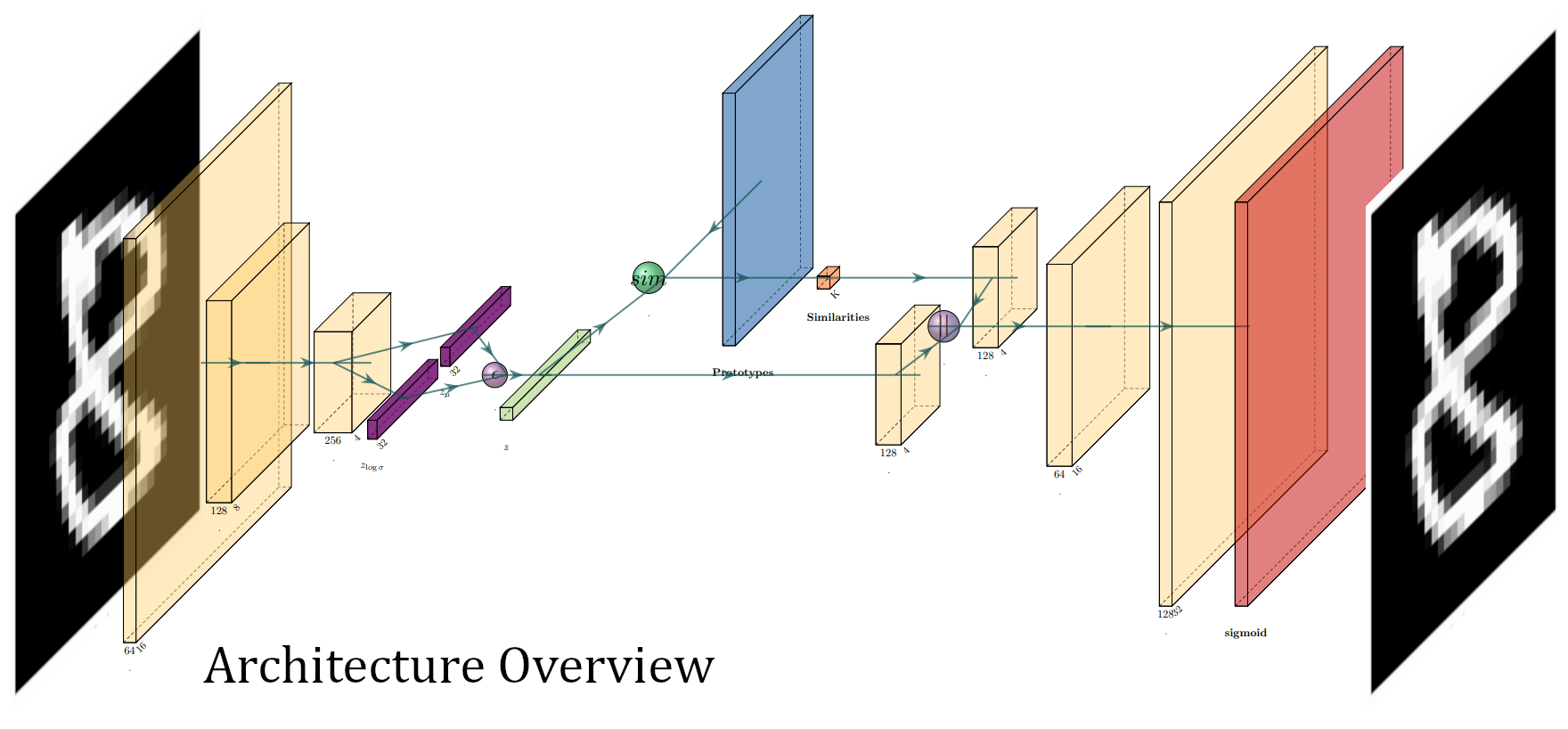}
    \caption{Overview of VAESim architecture. We use  $\epsilon$ to denote sampling with the reparametrization trick, $sim$ denotes the similarity measure, and $\parallel$ 1- the concatenation operation.}
    label{fig:overview}
\end{figure}

\begin{figure}[h]
    \centering
    \includegraphics[width=0.6\linewidth]{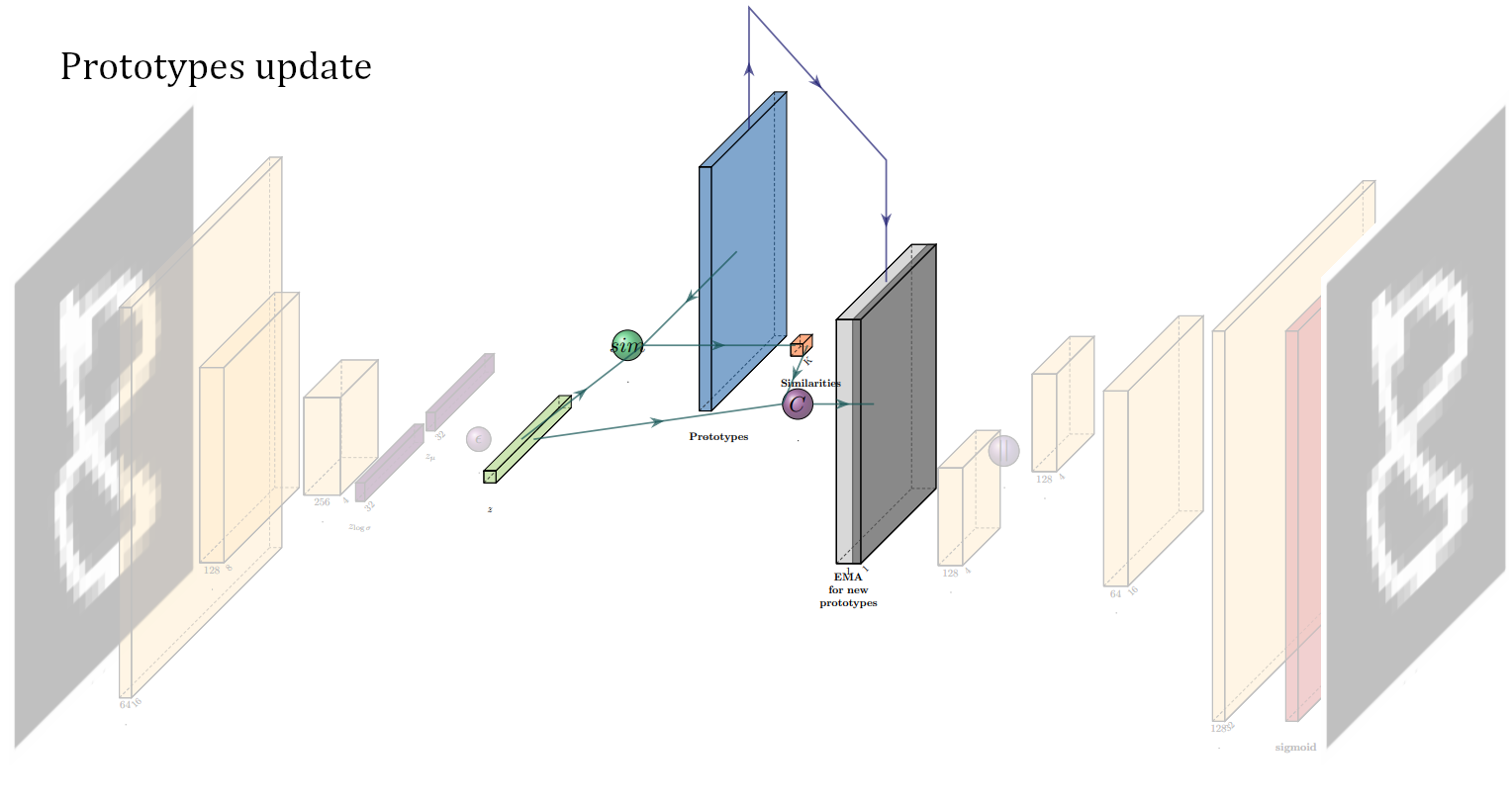}
    \caption{
    Graphical description of the $Q$ matrix update mechanism. See Algorithm \ref{alg:update} for details. We use the similarity between prototypes and sample embeddings to assign each sample embeddings to assign each sample embedding to the cluster with the most similar prototype vector. Then, we compute the mean of all $z$ embeddings in the same cluster and use the exponential moving average between them to update the prototypes matrix.}
    \label{fig:overview}
\end{figure}

The main goal of VAESim is to augment the vanilla VAE by The main goal of VAESim is to augment the vanilla VAE by conditioning the decoder not only on the latent embedding of a sample but also on its soft cluster assignment in latent space. This involves two main challenges:  learning a descriptive latent space that contains sufficient information to reconstruct the input samples and 2) learning the prototypical vectors of the latent space. The VAESim model is composed of three elements: the encoder  $e$, the decoder $d$, and the prototype matrix $Q^{K\times q}$. 

First, we encode the $p$-dimensional vector input $x$ with a neural network $e: \mathbb{R}^p \rightarrow  \mathbb{R}^q\times\mathbb{R}^q $ into a $q$-dimensional latent space $z$. Then, we sample a latent representation from the estimated posterior, modelled as a Gaussian, $z \sim \mathcal{N}(z_{\mu},e^{\frac{z_{\log{\sigma}}}{2}})$.
%The encoder is a neural network $e: \mathbb{R}^p \rightarrow  \mathbb{R}^q\times\mathbb{R}^q $ which maps the input sample $x$ from the input space with dimension $p$ into the latent space, so $z_{\mu},z_{\log{\sigma}}=e(x)$ of dimension $q\ll p$. We model the distribution over $z$ as a Gaussian and sample from the estimated posterior $z \sim \mathcal{N}(z_{\mu},e^{\frac{z_{\log{\sigma}}}{2}})$. 
We then compute the similarity between the latent embedding and the prototypical vectors using cosine similarity, and perform soft cluster assignment by computing the tempered softmax over the similarity measures.

\begin{equation}
    \label{eqn:cluster function}
	c = softmax(\frac{Qz}{\tau})
    %c=\frac{e^{\frac{(Qz)_i}{\tau}}}{\sum_i e^{\frac{(Qz)_i}{\tau}}}
\end{equation}

We use $\tau$ to denote the temperature parameter and $c$ to denote the cluster assignment. The latent representation $z$ is passed together with the cluster assignment $c$ to the decoder $d$ which maps back $\tilde{x}=d(z,c)$ into the input space to reconstruct the original input sample. We augment the standard ELBO loss of the vanilla VAE with an optional term that enforces orthogonality in the prototype representations.

The prototype vectors are updated following Algorithm \ref{algorithm}. The differentiating steps of our method are the \textit{cluster} equation and the \textit{update} function. The \textit{cluster} equation Eq. \ref{eqn:cluster function} produces soft assignments given the latent embeddings $z$ and the prototypes $Q^{K\times q}$. The \textit{update} function presented in Algorithm \ref{alg:update} is one of the core building blocks of our approach. At each training iteration, we assign all samples in the training dataset to a cluster. Each prototypical vector is updated with an exponential moving average with the mean of all sample embeddings assigned to its respective cluster. 

\begin{algorithm}[H]
  \caption{Training VAESim model}
  \label{algorithm}
  \begin{algorithmic}
    \INPUT{A set of N training images $X^{N\times p}$.}
    \STATE $\tau: 1.$ linearly decreases to 0.01 in half of the epochs.
    \STATE Initialize prototypes matrix $Q$.
    \FOR{$x$ in the training set}
     \STATE optimizer.zero\_grad()
      \STATE $z=e(x)$
      \STATE $c=softmax(\frac{Qz}{\tau})$
      \STATE $\tilde{x}=d(x,c)$
      \STATE loss=$\mathcal{L}_{recon}(x,\tilde{x})+\beta KL_{div}(q_z,p_z)$ 
      \STATE loss.backward()
      \STATE optim.step() 
      \STATE Update(Q,z,c)
    \ENDFOR
  \end{algorithmic}
\end{algorithm}

\begin{algorithm}[H]
  \caption{\textit{Update} function}
  \label{alg:update}
  \begin{algorithmic}
    \INPUT{$Q$,$z$,$c$}
    \STATE $\eta: 0.95$
    \FOR{$z_i$,$c_i$ in z,c}
     \STATE Sample cluster label $k_i \in [0,K-1 ]$: $k_i \sim multinomial(c_i)$.
    \ENDFOR
    
    \FOR{$i=0\dots K$}
     \STATE  Update the i-th column of $Q$ matrix using vectors assigned to it
     \STATE  $\bar{z_{i}}$=mean($z_i$) \textbf{if} $k_i==i$
     \STATE $Q_i=\eta \bar{z_{i}} +(1-\eta)Q_i$
    \ENDFOR
  \end{algorithmic}
\end{algorithm}

\subsection{VAE}
The main advantage of a variational autoencoder compared to standard autoencoders is that we impose a well-structured and continuous latent space, allowing us to sample meaningful examples from the latent space. Instead of simply projecting the data points into a manifold and then reconstructing them, in variational autoencoders the encoder neural network tries to approximate the joint probability $p(z|x)$ (which is computationally intractable) through probabilistic encoding $q_z(z|x)$. The encoder outputs the mean and the log-variance of a multidimensional Gaussian distribution. From that distribution, a value is sampled and passed to the decoder which tries to map back from $z$ to $\tilde{x}$.
The model is optimized by enforcing at the same time the quality of reconstruction through an MSE loss and a regularizer term that forces closeness of the Gaussian distribution to a chosen posterior which is a standard normal distribution. The latter term is usually a weighted Kullback-Leibler  (KL) divergence.
Therefore, the model is optimized to reconstruct items that have a latent distribution that resembles as much as possible a standard multivariate Gaussian distribution. 
The loss function is therefore $\mathcal{L}=\|x-\tilde{x}\|^2+\beta KL(q_z,\mathcal{N}(0,1))$.
Also, in this work, we modify the decoder network to take as input both the latent representation and a condition vector that is a measure of similarity across prototypes for each sample.

\subsection{Prototype Matrix Q}
When the model is initialized, the required number of clusters $k$ and the dimension of the latent space $p$ are initialized \textit{a priori}. The $Q$ matrix  should be a matrix with shape ($k$, $p$) having $k$ rows, each one with $p$ dimensions.
At the first forward pass, $k$ elements from the first batch are chosen randomly and set as rows of the $Q$ matrix.
In the early phase of training, the values of the prototypes are not meaningful so we use a tempered softmax of the similarity with a temperature schedule. For the first quarter of the training steps, the temperature is set to a high value ( $>1$) flattening the distribution of similarity across different prototypes. While the model’s ability to reconstruct images increases, the temperature decrease, results in harder conditioning vectors.
The key point related to the prototype matrix $Q$ is the way it is updated. Instead of relying on gradient descent, the values of the $Q$ matrix are computed separately from  the computational graph by using a momentum update of all samples belonging to a specific cluster, i.e.s all the samples whose z is most similar to a specific column of the $Q$ matrix.

At each iteration, the batch ``cluster labels'' are sampled from (or obtained by argmaxing) the categorical distribution $c$. These labels are used to compute the \textit{update} function for the $Q$ matrix.

\subsection{Evaluation}
The regularization constraint imposed by the KL divergence in the loss function tends to collapse all inputs as close as possible in the latent space. Thus, using classical metrics for unsupervised clustering could be misleading due to the commonly used normalization by the intercluster distance. Instead, we produce a visually plausible representation in 2D using t-SNE to observe how inputs are arranged in the latent space. Successively, we proposed three approaches for performance evaluation, all based on the downstream classification task. model training is therefore fully self-supervised way, and some of the available labels are used for comparison in the evaluation phase. We apply three standard methods for learning the downstream task: a statistical mapping approach between cluster labels and real labels, the kNN (k-nearest neighbors) algorithm and the training of a linear layer with categorical cross-entropy as the loss function.
In the statistical mapping approach, clusters are defined by argmaxing the similarity between samples and prototypes. Each cluster label is then associated with the most frequent label present in the related cluster, as a a way to quantify how well each prototype captures the key elements of a class. In inference, this mapping is used to predict the final class.
In the second approach, a subset of the training set is used to compute the "memory bank" for the kNN algorithm. In inference, each sample of the test set is compared with each element stored in the memory bank by computing the pairwise  euclidian distance  . The predicted label is defined as the mode of the label of the $k$ closest samples.

In the third approach, a linear layer maps from $z$ to the number of possible classes. This linear layer is trained with the Adam optimizer and a learning rate of $3*10^{-4}$ for 200 epochs over a subset of the training set, with categorical cross-entropy as the loss function.

\subsection{Baselines}
 In order to compare our probabilistic framework to classical two-step approaches, we trained a VAE build to be as similar as possible VAE we use to reconstruct images. Then, we performed a KMeans evaluation to produce cluster labels and train kNN and a linear classifier on the latent space as well as in other experiments. In this way VAE+KMeans can be evaluated on the same metrics used for our approach.   We also evaluated two other approaches, namely VaDE \cite{vade} and GMVAE \cite{gmvae} using default parameters. Each experiment was run $10$ times to obtain mean accuracies as well as standard deviations. It shoul dbe noted that reported baseline accuracies may not overlap with those shown in original papers because of different splits in the training, random seeds and, in particular, different ways to map between cluster labels and ground truth. Specifically, both GMVAE and VaDE rely on a measure of clustering accuracy defined by mapping $N$ cluster labels to $N$ real labels using the Hungarian matching algorithm \cite{hungarian}, which therefore requires the number of clusters to be indentical to the number of distinct labels. We choose to use a different mapping approach between cluster and actual labels to allow more freedom in the choice of cluster numerosity, hence allowing  the potential discovery of finer structures while maintaining the possibility of many-to-one mapping.
As comparison, we also report state-of-art results achieavable with transfer learning from ResNet50 pretrained on ImageNet.

\subsection{Experiments}

We tested our model on two datasets: a classical dataset of handwritten digits (\cite{deng2012mnist}), and a medical benchmarking dataset, PneumoniaMNIST, part of the MedMNIST dataset \cite{medmnist,medmnistv2}, which is a collection of medical imaging datasets arranged for benchmarking.
Pneumonia was chosen as a biomedical example, where chest X-ray images are assigned two possible labels: "healthy" and "pneumonia". This dataset is composed of 5856 images (training test split: 4708/1148). All code is written in PyTorch [20] and trained on a server with an A6000 GPU and 512 GB RAM.

Code for VAEsim implementation and other algorithms used as baselines can be found at \url{https://github.com/matteoferrante/VAESIM.git}

\section{Results}
Our main hypothesis is that our end-to-end deep clustering approach would improve performance in downstream classification tasks as compared to a two-step approach as well as other baselines. For the VAE+KMeans baseline, we train a VAE with architecture as close as possible to the ones used in our approach ( \textit{latent dimension} $32$ and \textit{batch size} $2048$).

Then, we used K-Means to evaluate the optimal number of clusters using the elbow method with the yellow-brick library \cite{yellowbrick}. VaDE and GMVAE were used as baselines for deep clustering. All results are described in `. Our model consistently outperformed (or performed on par in all metrics) within the baselines investigated, with improvements over the two-step approach ranging from $+13\%$ up to $+17\%$ on the MNIST dataset and from $+4\%$ to $+17\%$ on the Pneumonia dataset. The most relevant improvements are on the kNN accuracy measure, suggesting that our modification to the VAE framework drives the model to learn a space better suited for kNN estimations, i.e. positioning similar examples closer to each other in the latent space as compared to a standard VAE+KMeans.

\begin{table}[]

\label{tab:results}
\resizebox{\textwidth}{!}{\begin{tabular}{llllll}
\hline
\multicolumn{1}{|l|}{\textbf{Approach}} & \multicolumn{1}{l|}{\textbf{dataset}} & \multicolumn{1}{l|}{\textbf{statistical acc}} & \multicolumn{1}{l|}{\textbf{knn acc}} & \multicolumn{1}{l|}{\textbf{linear acc}} & \multicolumn{1}{l|}{\textbf{\begin{tabular}[c]{@{}l@{}}resnet50 acc\\ (supervised)\end{tabular}}} \\ \hline
VAE $\pm$ KMeans                              & MNIST                                 & 0.68 $\pm$ 0.01                                     & 0.80 $\pm$ 0.004                            & 0.68 $\pm$ 0.04                                & 0.99                                                                                              \\[2pt]
GMVAE                                   & MNIST                                 & 0.76 $\pm$ 0.01                                     & 0.91 $\pm$ 0.001                            & 0.76 $\pm$ 0.01                                & 0.99                                                                                              \\[2pt]
VaDE                                    & MNIST                                 & 0.81 $\pm$ 0.05                                     & 0.964 $\pm$ 0.001                           & \textbf{0.83 $\pm$ 0.008}                      & 0.99                                                                                              \\[2pt]
\textbf{VAEsim}                           & MNIST                                 & \textbf{0.83 $\pm$ 0.01}                            & \textbf{0.970 $\pm$ 0.001}                  & 0.81 $\pm$ 0.003                               & 0.99                                                                                              \\[2pt] \hline
VAE $\pm$ KMeans                              & Pneumonia                             & 0.631 $\pm$ 0.01                                    & 0.613 $\pm$ 0.014                           & 0.451 $\pm$ 0.033                              & 0.85                                                                                              \\[2pt]
GMVAE                                   & Pneumonia                             & 0.625 $\pm$ 0.001                                   & 0.627 $\pm$ 0.041                           & 0.429 $\pm$ 0.061                              & 0.85                                                                                              \\[2pt]
VaDE                                    & Pneumonia                             & 0.625 $\pm$ 0.001                                   & 0.747 $\pm$ 0.012                           & 0.590 $\pm$ 0.116                              & 0.85                                                                                              \\[2pt]
\textbf{VAEsim}                           & Pneumonia                             & \textbf{0.677 $\pm$ 0.001}                          & \textbf{0.778 $\pm$ 0.024}                  & \textbf{0.671 $\pm$ 0.010}                     & 0.85           \\[2pt]                                                                                  
\end{tabular}}

\vspace{2mm}

\caption{
Results on MNIST and PneumoniaMNIST datasets. The results are compared to different baselines and evaluated with accuracy measured with the statistical mapping approach to evaluate overlap between cluster labels and real labels, kNN to measure the closeness between similar samples in the latent space, and the classification done with a linear layer. The last column refers to the accuracy of a fine-tuned model from ResNet50 pretrained on ImageNet.}
\end{table}

\begin{figure}[h]
    \centering
    \includegraphics[width=0.6\linewidth]{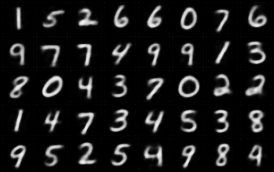}
    \caption{Prototypes for MNIST. These are the reconstruction of the $Q$ matrix rows.}
    \label{fig:mnistprototypes}
\end{figure}

\begin{figure}
    \centering
    \includegraphics[width=0.8\linewidth]{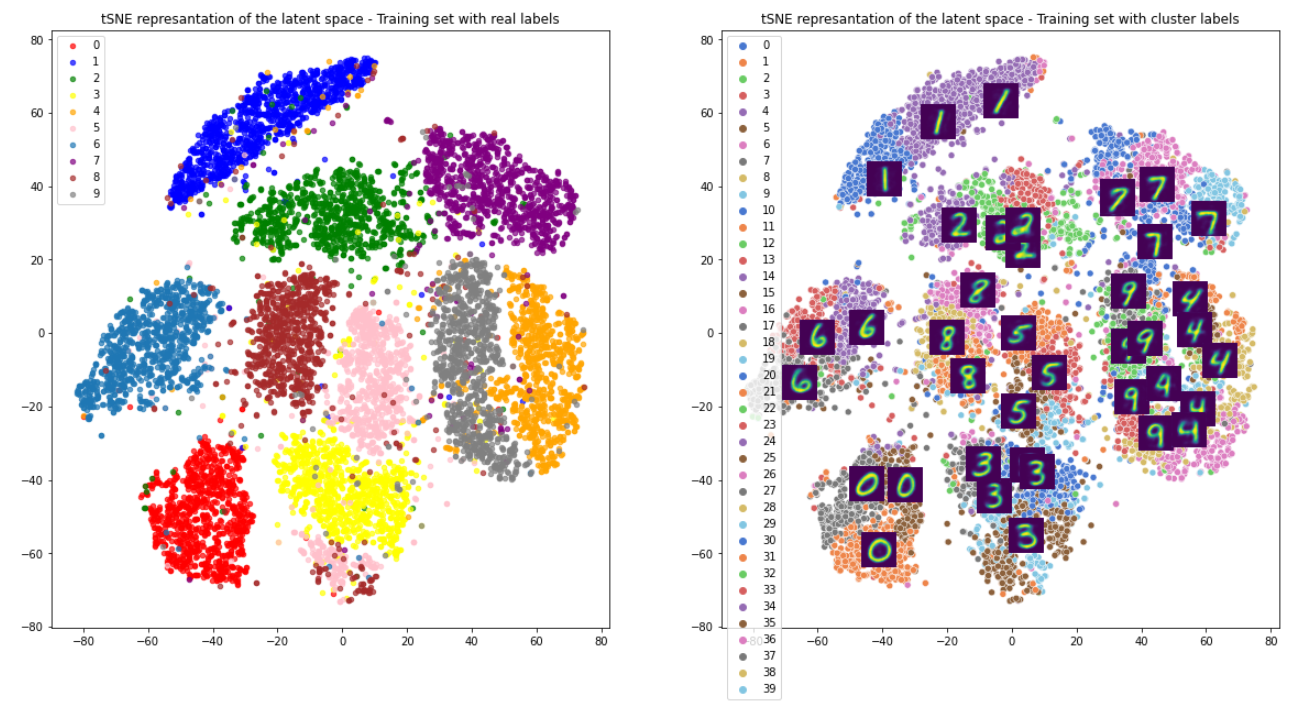}
  \caption{tSNE visualization of the latent space. Left: colors are real labels. Right: colors are cluster labels, with superimposed cluster prototypes.}\label{fig:tsnemnist}
\end{figure}

Examples of the \textit{prototypes} are visible in Fig \ref{fig:mnistprototypes}. Qualitatively, they clearly represent different ways to write digits and this pattern emerges from the update function during training. The  Fig \ref{fig:tsnemnist} shows a tSNE representation of the latent space with both the real labels and the cluster labels.

\begin{figure}[h]
    \centering
    \includegraphics[width=0.7\textwidth]{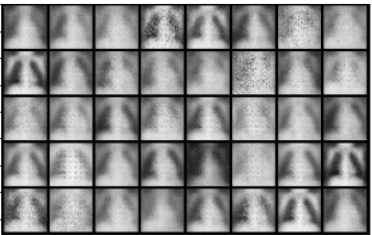}
    \caption{Prototypes for Pneumonia. These are the reconstruction of the $Q$ matrix rows.}
    \label{fig:pneumoniaprototypes}
\end{figure}

\begin{figure}
    \centering
    \includegraphics[width=0.7\textwidth]{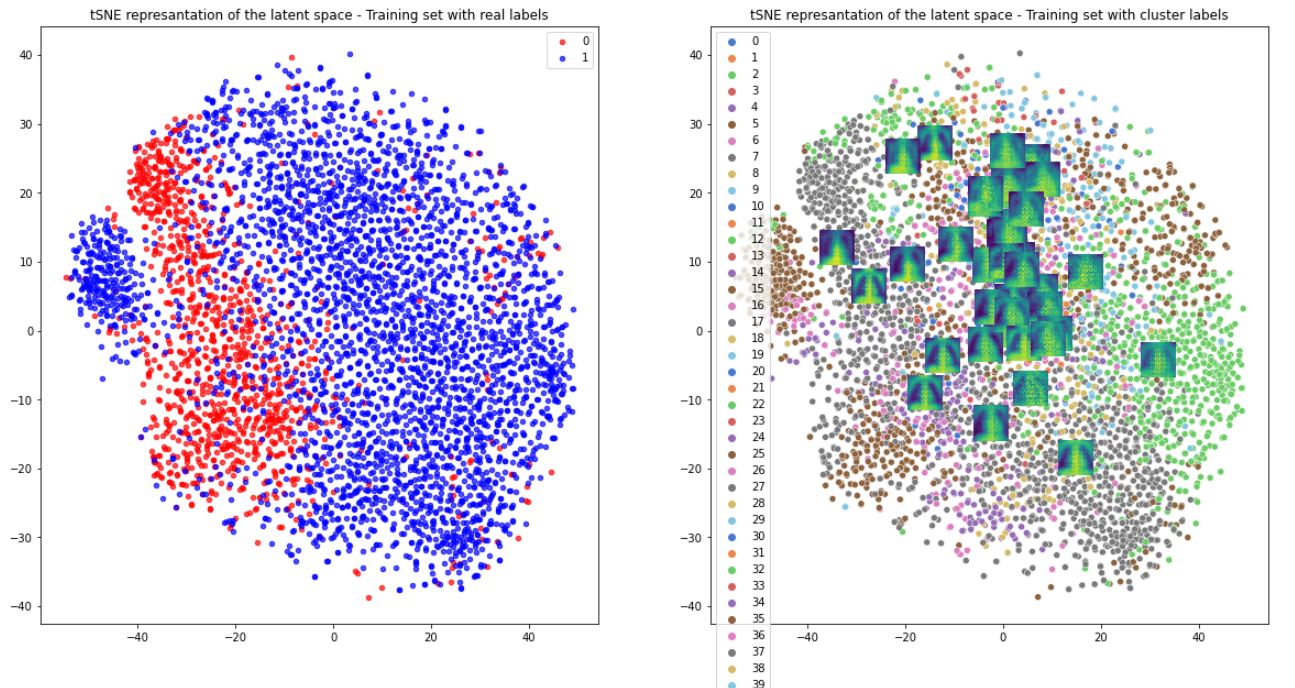}
  \caption{tSNE visualization of the latent space. Left: colors are real labels. Right: colors are cluster labels, with superimposed cluster prototypes.}\label{fig:tsnepneumonia}
\end{figure}

Prototypes and t-SNE representations for Pneumonia datasets are visible in the Fig \ref{fig:tsnepneumonia} and Fig \ref{fig:pneumoniaprototypes}.
The prototype representations include both some artefactual structures and some realistic X-ray chest images, where in many images pneumonia-related opacities are visible.

\section{Discussion}

In this work, we proposed a probabilistic framework that modifies the standard training of a variational autoencoder to generate representations of prototypes or modes present in the dataset. After encoding, the representations are compared with a memory bank of those prototypes, and a measure of similarity is passed to a conditional decoder, together with the latent representation, to generate reconstructions. The way we update the values of the prototypes is detached from the computational graph, allowing us to aggregate and update the information using nondifferentiable functions. As proof of concept, we choose an exponential moving average of the closest samples to the cluster prototype. At each step, the new value is computed from the actual value and average of all the samples in batch size for which this prototype is at a minimum distance.
In our approach, in the first part of the training, the focus of the encoder is to indirectly generate robust prototypes. It is an indirect process because the update of Q relies only on how close and similar same-class labels are in the latent space. Forcing the decoder to deal with the conditioning vector and using this updating procedure, we enforce an organization in the latent space that is better than the ones resulting from standard variational autoencoders. During the training process, the temperature  of the similarity conditioning vector decreases, resulting in harder assignments. When the prototypes became more stable, the encoder focuses more on exploiting features in terms of differences between samples and prototypes. We showed that our continuous conditional learning with decreasing temperature improves the latent space organization with representation, which is also useful for the downstream classification task and results in better performances as compared to two-step approaches. Also, our approach outperforms the baselines of other approaches based on deep clustering variational autoencoders. All other VAE proprieties are unchanged, hence allowing to sample from the latent space and reject bad samples based on their similarity with a specific cluster. For example, looking at cluster prototypes obtained for the model trained on MNIST, one could select clusters that represent a digit or a specific way to write a digit and sample from the latent space random values, possibly keeping only the ones that surpass a specific threshold of similarity with that cluster, to reconstruct similar digits. In the same way, if able to interpret prototypes for X-ray imaging or other medical imaging, one could reconstruct specific modes of healthy or pathological images to investigate their proprieties. Importantly, our model is not targeted to maximum accuracy because it is trained for a different objective (image reconstruction). Still, as an emergent property, we obtain a well-organized latent space that can also be exploited for a downstream task. Our model can perform simultaneous reconstruction and clustering. This could be useful, for example, in guided data augmentation or in patient stratification beyond what is visible to the naked eye of the clinician.

It should be noted that several hyperparameters need initialization. In particular, the most challenging one is the number of prototypes. In  \ref{sec:appendix} An investigation of the hyperparameter effect is presented, where we acknowledge that the computational efforts necessary to find the best hyperparameters for any given task could represent one of the main limitations of this model. Imposing a high number of prototypes may improve performance, however, each cluster contains a sufficient number of samples, hence requiring a larger batch size. Finding the optimal way to initialize the Q matrix in terms of values and dimensions will be the object of future investigations.

\section{Conclusion}
 In this work, we proposed a probabilistic framework that improves latent space generalizability for reconstruction and downstream tasks by learning how to reconstruct images through latent representation coupled with a measure of similarity with prototypes that represent the modes present in the dataset. We dynamically update representations during the training, forcing the model to implicitly take into account features that enhance similarities and differences between samples and prototypes. We demonstrate that our approach is better than a two-step approach basedon the combination of a VAE and KMeans, and training these models in an end-to-end fashion could lead to up $+10\%$ in performance on average. We also outperformed baselines based on different modifications of the VAE framework for deep clustering on most of the metrics that we defined. Also, our model produces a latent space that generalizes well when used for the downstream task. We evaluated performance by examining the ability of cluster labels to recall specific classes, the organization in latent space probed by the use kNN algorithm for classification and its separability using a linear classifier. We also reached performances similar to the ones obtained fine-tuning very large models in a supervised way by starting with a self-supervised pretraining of our modified VAE.

\bibliographystyle{splncs04}
\bibliography{samplepaper}

\newpage
\section*{Appendix: Hyperparameters impact}

\begin{figure}[ht]
    \centering
    \includegraphics[width=0.75\textwidth]{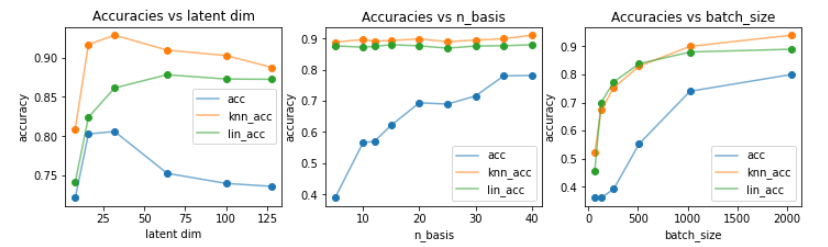}
    \caption{Comparison between different metrics across values of \textit{latent dim} (Left), \textit{number of prototypes} (Center) and \textit{batch size} (Right) for the MNIST dataset.
    Blue line report the accuracy computed with the statistical classification, the orange line the accuracy obtained with the kNN approach while the green line is related to the classification obtained with the linear classification.}

    \label{fig:metricscomparisonmnist}
\end{figure}

This section explores how different hyperparameters impact on the performances of our model.
In particular, the latent dimension, the number of prototypes and the batch size are the most important things that describe the model. We evaluated all this by performing a sweep over different values of the hyperparameters, changing just one at time keeping all the configurations fixed. We used the Weight \& Biases library for this scope \cite{wandb}.
For all these approaches we kept a fixed architecture, composed by an encoder composed by three convolutional layers with \textit{relu} activation and batch normalization, followed by flattening and dense linear layers to output the $z_{\mu}$ and $z_{\log{\sigma}}$.
Convolutions layers have stride=2, kernel size=4 and padding=1.
The $Q$ matrix of prototypes is just a Pytorch matrix, it doesn't require to be a learnable parameter because its evolution is detached from the computational graph to make it possible using different types of function to update its values.
The decoder is almost symmetrical to the encoder, using transposed convolutions instead of standard convolutions. Being a conditional decoder there is a first linear layer that maps the conditioning vector as well as the latent vector to the features of the first convolutional layer by concatenating them.
There's also an output convolutional layer with as many channels as the input image (1 for grayscale images and 3 for rgb images) with \textit{sigmoid} activation.
In Fig \ref{fig:metricscomparisonmnist} the impact of latent dimension, batch size and number of prototypes on classification accuracy is shown as a characterization study over the MNIST dataset.
The statistical accuracy, kNN accuracy and linear accuracy are reported as functions of the investigated parameter.

For \textit{latent dimensions} it seems to be better to use a value around $32$, where the statistical accuracy and the kNN accuracy present a peak and the linear accuracy starts to converge to the optimal value.
Increasing the \textit{number of prototypes} (referred in the figure as n basis) is beneficial for the statistical accuracy (because there are more clusters to choose and this probably reduces the variance inside each assignment), while there is almost no influence for kNN and linear accuracy.
Increasing \textit{batch size} has a beneficial effect for all the metrics.

\end{document}